\documentclass[10pt,twocolumn,letterpaper]{article}

\usepackage[pagenumbers]{cvpr} 
\usepackage{graphicx}
\usepackage{amsmath}
\usepackage{amssymb}
\usepackage{booktabs}
\usepackage{pifont} 
\usepackage{colortbl}
\usepackage{xcolor}
\usepackage{algorithm}
\usepackage{algpseudocode}
\usepackage{makecell}
\usepackage{multirow}
\usepackage{bm}
\usepackage{enumitem}
\usepackage{newtxtext,newtxmath}
\usepackage{algorithm}
\usepackage{algpseudocode}
\usepackage{amsmath}

\definecolor{cvprblue}{rgb}{0.21,0.49,0.74}
\usepackage[pagebackref,breaklinks,colorlinks,allcolors=cvprblue]{hyperref}

\def\paperID{46349} 
\def\confName{CVPR}
\def\confYear{2026}

\title{RoadGIE: Towards A Global-Scale Aerial Benchmark for Generalizable Interactive Road Extraction}

\author{
Chenxu Peng$^{2,1}$,
Chenxu Wang$^{2,1}$,
Yimian Dai$^{1,2,3}$,
Yongxiang Liu$^{4}$,
Ming-Ming Cheng$^{1,2,3}$,
Xiang Li$^{1,2,3}$\thanks{Corresponding author: Xiang Li.}\\
$^1$ NKIARI, Shenzhen Futian
$^2$ VCIP, CS, Nankai University
$^3$ AAIS, Nankai University\\
$^4$ College of Electronic Engineering, National University of Defense Technology, Changsha, China\\
{\tt\small cxpeng@mail.nankai.edu.cn},
{\tt\small facias914@gmail.com},
{\tt\small yimian.dai@gmail.com},\\
{\tt\small lyx\_bible@sina.com}, 
{\tt\small \{cmm, xiang.li.implus\}@nankai.edu.cn}
}

\makeatletter
\let\@oldmaketitle\@maketitle
\renewcommand{\@maketitle}{\@oldmaketitle
  \begin{center}
    \centering
    \includegraphics[width=0.95\textwidth]{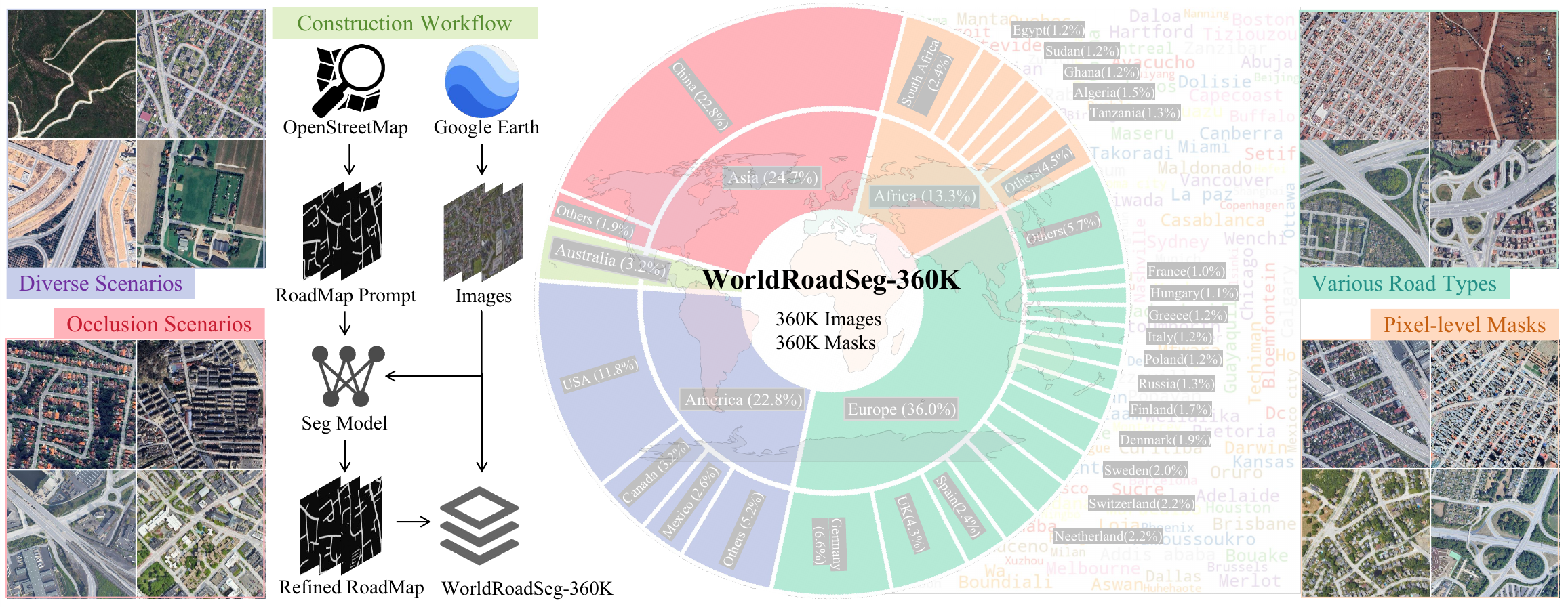}
    \captionof{figure}{Composition of the proposed dataset. We visualize the collection workflow and geographical distribution of the \texttt{WorldRoadSeg-360K} dataset. Furthermore, representative examples of road segmentation across typical terrain types are displayed on both sides of the figure.}
    \label{fig:proposed_dataset}
  \end{center}
  \vspace{0.5em} 
}
\makeatother

\begin{document}

\maketitle


\begin{abstract}
   Accurate road segmentation from aerial imagery is fundamental to many geospatial applications. However, existing datasets often suffer from limited scene diversity, low semantic granularity, and poor structural continuity, restricting their generalization across environments. To address these challenges, we introduce \texttt{WorldRoadSeg-360K}, the largest and most diverse road segmentation dataset to date, comprising 366,947 high-resolution images collected from 38 countries and 223 cities across various terrains and continents. \texttt{WorldRoadSeg-360K} serves as a comprehensive benchmark and reveals key challenges in handling diverse and structurally complex scenes. Automated approaches often struggle to preserve road connectivity, while current interactive methods lack efficient, topology-sensitive tools for real-world road editing. To this end, we present \texttt{RoadGIE}, establishing a novel interactive paradigm for road extraction in remote sensing. Unlike prior point- or box-based prompting strategies, \texttt{RoadGIE} supports connectivity-aware prompts, including clicks and scribbles, which inherently align with the topology of road networks. To improve structural consistency and mitigate performance degradation during iterative interactions, \texttt{RoadGIE} integrates an expert-guided prompting strategy and adapts the skeleton-based recall loss for interactive scenarios. \texttt{RoadGIE} achieves state-of-the-art performance in both segmentation accuracy and topological consistency on \texttt{WorldRoadSeg-360K} and other benchmarks, while maintaining efficient operation with only 3.7M parameters. The code are publicly available at: \url{https://github.com/chaineypung/RoadGIE}.
\end{abstract}    
\begin{figure}[t]
    \vspace{-1em}
    \centering
    \includegraphics[width=0.45\textwidth]{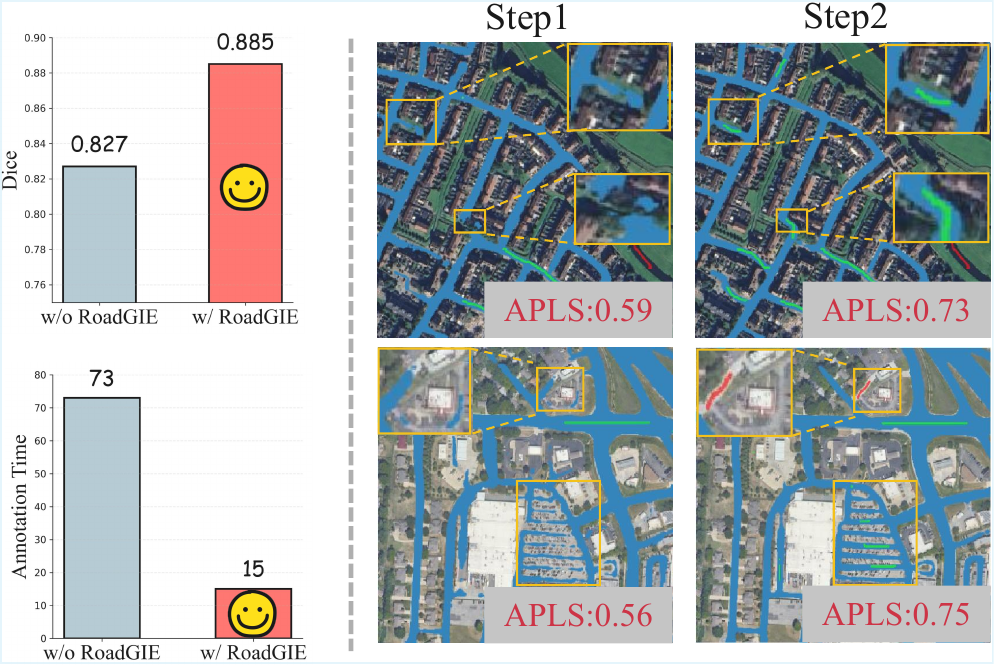}
    \caption{The bar charts on the left compare the average annotation accuracy and time consumption for labeling 100 images, with and without \texttt{RoadGIE}. To evaluate its generalization ability, we visualize the segmentation results and step-by-step user interactions on previously unseen datasets, as shown on the right. Throughout the interaction process, positive inputs are indicated in \textcolor{green}{green}, negative ones in \textcolor{red}{red}, and the model's predicted masks are displayed in \textcolor{blue}{blue}.}
    \label{fig:pipeline}
    \vspace{-1em}
\end{figure}

\vspace{-3.2\baselineskip}   
\nopagebreak 
\section{Introduction}
Accurate identification of road regions in remote sensing imagery is fundamental for road modeling~\cite{wang2016review, wu2025ret3d, gao2025towards}, map updating~\cite{chen2024updating}, and spatial structure analysis~\cite{wu2014assessing}. As a key technology supporting spatial element extraction and geographic information system construction~\cite{mena2005automatic}, road segmentation has become a core task in geospatial visual understanding. However, due to the inherent complexity of remote sensing imagery, achieving high-precision road segmentation heavily depends on the diversity of the datasets.

Existing road datasets struggle to simultaneously balance scene diversity, semantic granularity, and structural continuity. For instance, while the Global-Scale dataset~\cite{yin2024towards} offers global coverage, its vectorized centerlines miss road width and boundary continuity, limiting its use in fine-grained segmentation and morphology modeling. In contrast, the LSRV dataset~\cite{lu2021gamsnet} provides high-precision pixel-level annotations but has limited samples and mainly urban coverage, lacking diverse terrains and complex morphology.

To address these, we introduce \texttt{WorldRoadSeg-360K}, as shown in Fig.~\ref{fig:proposed_dataset}, the first globally distributed road segmentation dataset, comprising 366,947 images collected across 38 countries and 223 cities, with spatial resolutions ranging from 0.8 to 1.1 meters,  following the standards set
by~\cite{he2020sat2graph}. It spans all continents except Antarctica and covers diverse environments, including urban, rural, and mountainous areas. To facilitate comprehensive algorithm evaluation, we further incorporate an out-of-domain (OOD) test set, consisting of all images from the LSRV dataset~\cite{lu2021gamsnet}, to more realistically assess cross-domain generalization capabilities.

While high-quality datasets provide a solid foundation for model training, the geometric complexity, topological continuity, and semantic ambiguity of road regions in remote sensing imagery impose higher demands on segmentation methods. In practical applications, purely relying on automated segmentation models often results in fragmented road maps, which are insufficient for high-precision road modeling and map generation~\cite{mei2021coanet}. Although the SAM family~\cite{kirillov2023segment, xiong2024efficientsam, zhao2023fast} of interactive models demonstrates strong generalization on natural images, their point- or box-based prompting strategies exhibit limited suitability for road segmentation tasks. Moreover, high latency and user intent ambiguity further degrade the user experience~\cite{wong2024scribbleprompt, ravi2024sam}. 

\begin{table*}[t]
\centering
\caption{
Comparison of different road extraction datasets. 
To ensure fair comparison despite differences in image quantity and scale, 
all images were uniformly divided into $512 \times 512$ patches, 
with patches containing no roads or mostly black backgrounds removed. U, R denote urban and rural areas, respectively.
}
\label{tab:dataset}
\definecolor{Gray}{gray}{0.9}
\scriptsize
\setlength{\tabcolsep}{4pt}
\renewcommand{\arraystretch}{1.05}
\resizebox{0.95\linewidth}{!}{
\begin{tabular}{cccccc@{\hspace{12pt}}cccccc}
\toprule
\textbf{Dataset} & \textbf{Res.} & \textbf{Images} & \textbf{Cities} & \textbf{Types} & \textbf{Anno.} &
\textbf{Dataset} & \textbf{Res.} & \textbf{Images} & \textbf{Cities} & \textbf{Types} & \textbf{Anno.} \\
\midrule
Massachusetts & 1 & 10,413 & $>$1 & U & Line &
LoveDA & 0.3 & 3,154 & 1 & U, R & Mask \\
RoadNet & 0.21 & 54 & 1 & U & Mask &
LSRV & 0.3--0.6 & 1,787 & 3 & U, R & Mask \\
RoadTracer & 0.6 & 9,600 & 40 & U & Line &
HUAWEI & 0.8 & 6,953 & 1 & U & Mask \\
SpaceNet & 0.3 & 11,120 & 4 & U & Line &
WHU-road & 1 & 4,798 & 1 & U & Mask \\
DeepGlobe & 0.5 & 24,904 & $>$3 & U, R & Mask &
Global-Scale & 1 & 55,488 & -- & U, R & Line \\
CHN6-CUG & 0.5 & 3,032 & 6 & U & Mask &
\textbf{WorldRoadSeg-360K} & 
\textbf{0.8--1.1} & 
\textbf{366,947} & 
\textbf{223} & 
\textbf{U, R} & 
\textbf{Mask} \\
\bottomrule
\end{tabular}
}
\end{table*}

In summary, the low accuracy of fully automatic methods, the incompatibility of existing interactive models with road-specific characteristics, the high latency of mainstream interactive segmentation frameworks, and the ambiguity in user intent pose significant challenges to interactive road extraction in remote sensing. Visual prompting strategies should align with the morphological characteristics of the target objects. Scribble-based prompts~\cite{li2024prism, isensee2025nninteractive}, by encoding connectivity and topological continuity, guide the model to capture fine-grained structural information with minimal interaction effort, which is particularly crucial for road segmentation.

Based on these insights, we propose \texttt{RoadGIE}, a real-time interactive road segmentation framework with the following features: 1) a compact 3.7M model with high efficiency for real-time inference, and 2) support for connectivity-aware prompts (clicks \& scribbles), as shown in Fig.~\ref{fig:pipeline}. To achieve these goals, we emphasize usability and realistic interaction in system design. First, we leverage the \texttt{WorldRoadSeg-360K} dataset for diverse, high-quality training samples. Second, we introduce an expert-guided prompting strategy to guide prompts to uncertain regions during training, thereby simulating realistic user interactions. Third, to alleviate user intent ambiguity and improve interaction stability, \texttt{RoadGIE} incorporates topo-semantic road instantiation during training, combining topology and semantics. Finally, to mitigate performance degradation during iterative interactions, we adapt the skeleton-based recall loss~\cite{kirchhoff2024skeleton} for interactive segmentation, helping the model preserve connectivity across rounds.

Our main contributions are summarized as follows:
\begin{itemize}[leftmargin=8pt, topsep=2pt, itemsep=1pt, parsep=0pt]
    \item \texttt{WorldRoadSeg-360K}, the largest road segmentation dataset in remote sensing, is constructed with 366,947 images from 223 cities across 38 countries.
    \item \texttt{RoadGIE} is introduced as the first generalizable interactive framework for road extraction, achieving state-of-the-art accuracy and connectivity on multiple benchmarks, while supporting real-time, cost-efficient annotation.
    \item A comprehensive interaction strategy is developed, incorporating an expert-guided prompting mechanism, a topo-semantic aware road instantiation module to reduce user intent ambiguity, and an adapted skeleton-based recall loss to preserve structural consistency across interactions.
\end{itemize}
\section{Related Work}

\subsection{Road Segmentation Datasets}

Over the past decade, many benchmark datasets for road extraction have been introduced~\cite{van2018spacenet, mnih2013machine, yin2024towards, demir2018deepglobe, wang2021loveda, lu2021gamsnet, bastani2018roadtracer}, significantly advancing research in this field. These datasets can be broadly divided into two types~\cite{yin2024towards}: graph-labeled datasets and segmentation-labeled datasets. Graph-labeled datasets represent roads as vector-based graphs. Early examples like SpaceNet~\cite{van2018spacenet} and Massachusetts~\cite{mnih2013machine} focus on cities such as Las Vegas, Shanghai, and Boston. To cover complex terrains like farmlands and mountains, the Global-Scale dataset~\cite{yin2024towards} was recently released. It achieves global coverage by leveraging road centerlines from OpenStreetMap (OSM)~\cite{haklay2008openstreetmap}. However, OSM quality varies across regions and often lacks road width details~\cite{xi2024pixels}, posing challenges for model training.

Segmentation-labeled datasets provide pixel-level road masks. Notable examples include DeepGlobe~\cite{demir2018deepglobe}, LoveDA~\cite{wang2021loveda}, and LSRV~\cite{lu2021gamsnet}, which cover cities such as Wuhan, Birmingham, and Shanghai. Nevertheless, most are limited to one country or city, lacking geographic diversity and hindering model generalization. In fact, pixel-level datasets are also more costly and labor-intensive to construct. This work constructs a large-scale, high-precision global road segmentation dataset as a universal benchmark for performance evaluation.


\subsection{Connectivity-based Segmentation}

\begin{figure*}[t!]
    \centering
    \includegraphics[width=1.0\textwidth]{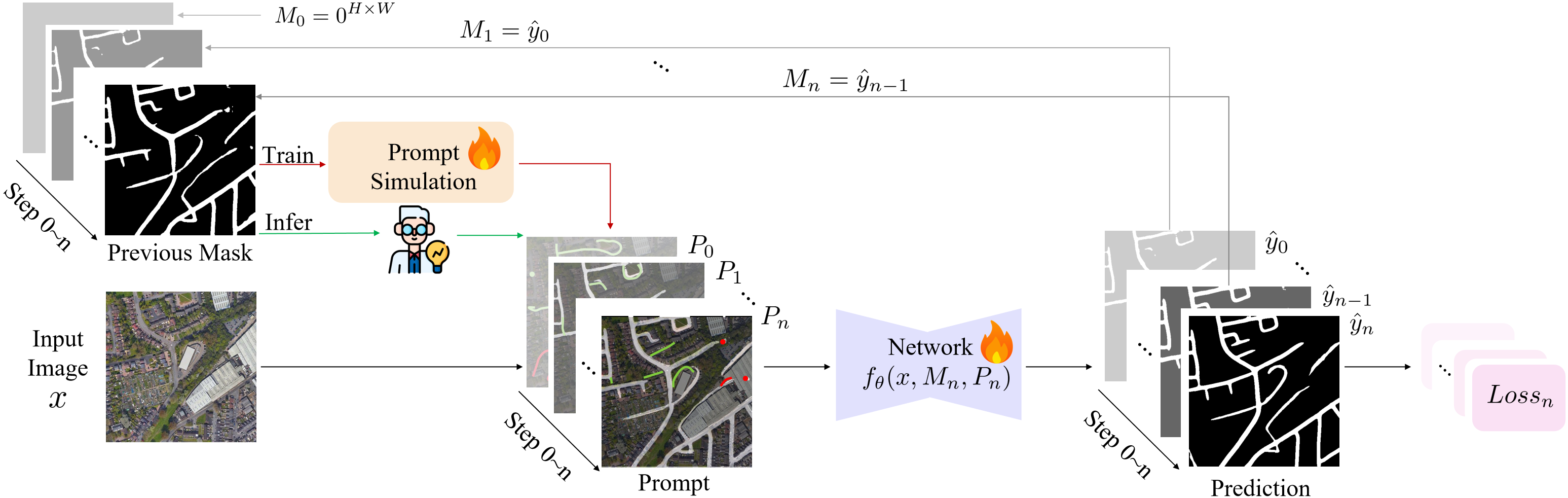} 
    \caption{Overview of our \texttt{RoadGIE} pipeline. Given an input image and an initial guidance prompt, the network produces a preliminary segmentation. This output is compared against the ground truth to compute training loss and localize misclassified regions. A simulated annotator then generates corrective hints in response to these errors. These updated prompts, combined with the model’s latest prediction, are fed back into the network for iterative refinement.}
    \label{fig:network}
\end{figure*}

Accurate road segmentation relies on preserving topological structures, such as layout and connectivity~\cite{qi2023dynamic, kampffmeyer2018connnet, chen2024connectivity}. Recent methods propose specialized architectures to model roads’ elongated and continuous nature. For example, strip convolutions~\cite{mei2021coanet} leverage shape priors to extract directional context, while connectivity-aware attention~\cite{song2024uroadnet} addresses prediction gaps caused by occlusions. Although effective, these approaches often overlook explicit modeling of pixel connectivity, which is critical for structural completeness.

Another direction focuses on learning connectivity directly during training~\cite{yang2023directional, zhang2024graphmorph, zhang2022progressive, shit2021cldice, viti2022coronary, menten2023skeletonization, peng2025simple}. Skeleton-based losses like clDice~\cite{shit2021cldice}, have shown promise in topology-aware segmentation. Viti et al.\cite{viti2022coronary} extended it with soft-persistent skeletons for vessel modeling, and Menten et al.\cite{menten2023skeletonization} proposed a more robust skeletonization algorithm, albeit with high computational cost. In this work, we integrate skeleton-based recall loss~\cite{kirchhoff2024skeleton} into interactive segmentation to mitigate performance degradation across user iterations.

\subsection{Interactive Segmentation}

Interactive segmentation is a human-in-the-loop framework where user inputs, such as clicks or boxes, iteratively guide models to refine object masks with minimal supervision. Recent advances in foundation models, such as the Segment Anything Model (SAM)\cite{kirillov2023segment}, have enabled class-agnostic segmentation across diverse domains using simple prompts, thereby reducing the need for task-specific tuning. Despite their effectiveness, these methods often struggle with objects that have ambiguous or fragmented boundaries~\cite{huang2024segment, mazurowski2023segment}. This limitation is particularly evident in remote sensing imagery~\cite{sultan2023geosam}, where linear structures like roads are frequently occluded by buildings, vegetation, or shadows. In such cases, point- or box-based prompts provide only coarse spatial cues and fail to convey detailed shape or structural continuity~\cite{wong2024scribbleprompt}.

To address these, scribble-based prompts have been proposed as a more expressive and intuitive alternative~\cite{chen2023scribbleseg, li2024prism, li2024scribformer, zhang2025exploiting, qiu2025sparsemamba}. Scribbles naturally encode object shape, continuity, and connectivity, and align well with how human annotators interact with visual data. Due to the limited availability of real scribble annotations, early studies explored synthetic scribble data~\cite{lee2020scribble2label} or simulated inputs based on simplified curves~\cite{agustsson2019interactive}. More recent frameworks, such as ScribblePrompt~\cite{wong2024scribbleprompt}, employ scribble simulation engines during training and achieve strong results on medical segmentation benchmarks. In this work, we argue that the design of visual prompts should reflect the morphological characteristics of target objects. Scribble inputs, by providing fine-grained structural cues, allow models to better capture the continuity of thin and occluded regions. This is particularly beneficial in remote sensing scenarios where accurate segmentation of roads requires connectivity priors.
\section{A Global-Scale Aerial Benchmark Dataset}
\label{dataset}

We conduct a comprehensive survey of existing publicly available road segmentation datasets, as summarized in Table~\ref{tab:dataset}. Among them, LSRV~\cite{lu2021gamsnet} comprises three distinct urban styles, each annotated with high-quality, pixel-level road masks, enabling fine-grained urban scene understanding. Global-Scale dataset~\cite{yin2024towards} represents the largest road vector dataset to date, capturing major road networks across metropolitan areas worldwide. These datasets address the limitations of earlier benchmarks such as Massachusetts~\cite{mnih2013machine} and DeepGlobe~\cite{demir2018deepglobe}, which are geographically constrained to a single city or region, thereby enabling broader generalization and cross-domain evaluation.

In this work, we introduce \texttt{WorldRoadSeg-360K}, a large-scale road segmentation dataset that spans all continents except Antarctica. Compared to prior datasets such as Global-Scale~\cite{yin2024towards}, \texttt{WorldRoadSeg-360K} offers significant advancements in terms of data volume, geographic diversity, and annotation quality. It is designed to address the shortcomings of current road extraction benchmarks and serve as a comprehensive resource for evaluating model performance in varied real-world scenarios. Our dataset construction process began with a global urban area survey, from which we systematically selected rectangular regions (15–45 km in size) representing cities of different scales. Each region is geo-located and includes a variety of terrain types, such as dense urban centers, rural outskirts, and mountainous areas. High resolution satellite imagery was obtained using the Google Static Maps API~\cite{svennerberg2010beginning}, while coarse road annotations were sourced from OSM. To improve label accuracy, as shown in Fig.~\ref{fig:proposed_dataset} these annotations were used as prompts for several state-of-the-art segmentation models, including SAM~\cite{kirillov2023segment}, HQ-SAM~\cite{ke2023segment}, and RobustSAM~\cite{chen2024robustsam}. The outputs from these models were fused and combined with the original annotations to produce refined road masks. A subsequent manual validation phase was carried out to categorize the dataset into high- and low-quality subsets, with the high-quality portion intended for fine-tuning during later stages of model training. 

In total, \texttt{WorldRoadSeg-360K} encompasses 223 cities across 38 countries, with a geographic coverage roughly four times larger than the Global-Scale dataset. It consists of 366,947 satellite images, each sized at 512×512 pixels with a spatial resolution of 0.8-1.1 meter per pixel, in accordance with the standard defined in~\cite{he2020sat2graph}. To evaluate model generalization to unseen domains, we additionally collected 1,789 images from Boston, Birmingham, and Shanghai. These images are excluded from the training set and designated as an OOD test set, enabling robust assessment across diverse geographic contexts.
\section{Generalizable Interactive Road Extraction}
\label{method}

We introduce \texttt{RoadGIE}, a novel interactive segmentation framework specifically designed to address the distinctive geometric properties of road networks, which are characterized by high aspect ratios, extended continuity, and strong topological sensitivity. As illustrated in Fig.~\ref{fig:network}, the framework operates in an iterative fashion and refinement strategy, progressively refining segmentation results through successive rounds of interactive user supervision. At each iteration $n$, the model takes as input the image $x$, the preceding prediction mask $M_n$, and a set of prompts $P_n$ provided either by the user or simulated agents. Based on these inputs, it produces an updated prediction $\hat{y}_n$. This refinement process continues for a predefined number of iterations or until the segmentation reaches a desired level of accuracy.

Formally, the segmentation at step $n$ is defined as:

\vspace{-1.2em}
\begin{equation}
    \hat{y}_n = f_\theta(x, M_n, P_n),
    \label{eq:step_forward}
\end{equation}
\vspace{-1.2em}

where $f_\theta$ denotes the segmentation network with parameters $\theta$, and $M_n = \hat{y}_{n-1}$ is the previous prediction (with $M_0 = 0^{H \times W}$). In the following, we describe the key components of \texttt{RoadGIE}: the network architecture, prompt simulation strategy, expert-guided prompt generation, and prompt-excluded skeleton loss.

\subsection{Network Architecture}

We adopt a lightweight UNet~\cite{ronneberger2015u} as the backbone to extract multi-scale features with high spatial resolution, suitable for real-time interactive segmentation. To enhance road structural continuity, we follow~\cite{yuan2025strip, hou2020strip, mei2021coanet} and introduce a Directional Aggregation Module (DAM) at the decoder output. It captures long-range dependencies along dominant directions through efficient 1D convolutions.

Let $F_n \in \mathbb{R}^{H \times W \times C}$ denote the decoder feature map at interaction step $n$. For each direction $D = (d_h, d_w) \in \mathcal{D}$, DAM extracts a directional response $Z_D \in \mathbb{R}^{H \times W \times C}$ by applying a 1D convolution along $D$:

\vspace{-1.2em}
\begin{equation}
Z_D[i, j, c] = \sum_{l=-k}^{k} F_n[i + l d_h, j + l d_w, c] \cdot w_D[k - l] + b_D,
\label{eq:strip_conv_per_direction_bias}
\end{equation}
\vspace{-1.2em}

where $w_D \in \mathbb{R}^{2k+1}$ is a learnable 1D kernel specific to direction $D$, and $b_D \in \mathbb{R}$ is the corresponding scalar bias term. The convolution is applied independently for each channel $c$, and the offset $l$ determines the sampling pattern along direction $D$. We define $\mathcal{D} = {(1,0), (0,1), (1,1), (-1,1)}$, corresponding to vertical, horizontal, diagonal, and anti-diagonal directions.

The outputs $\{Z_D\}$ are concatenated and passed through a $1 \times 1$ convolution, followed by a sigmoid activation:

\vspace{-1.2em}
\begin{equation}
\hat{y}_n = \sigma _{>0.5}\left( \text{Conv}_{1\times1} \left( \operatorname{concat}_{D \in \mathcal{D}} \left[ Z_D \right] \right) \right),
\label{eq:scm_output}
\end{equation}
\vspace{-1.2em}

where $\operatorname{concat}$ denotes channel-wise concatenation, and $\sigma(\cdot)$ is the element-wise sigmoid function. The final output $\hat{y}_n$ represents the predicted binary mask at step $n$. DAM enhances connectivity-aware predictions by promoting directional consistency across road-like structures.

\subsection{Prompt Simulation}
Interactive segmentation relies on user inputs such as clicks or scribbles to iteratively refine predictions. However, collecting large-scale manual prompts during training is impractical. To overcome this, we employ prompt simulation, which synthetically generates user-like corrections based on prediction errors. This allows the model to learn to incorporate interactive feedback without requiring real user annotations. During training, we simulate user interactions by identifying regions where the model's prediction deviates from the ground truth. At each interaction step $n$, we compute the error map $\varepsilon_n = y - \hat{y}_{n-1}$ and sample a correction region $V \subset \varepsilon_n$ as the basis for prompt generation. We employ two types of synthetic prompts: point and scribble interactions.

\begin{figure}[t]
    \centering
    \vspace{-8pt} 
    \includegraphics[width=0.43\textwidth]{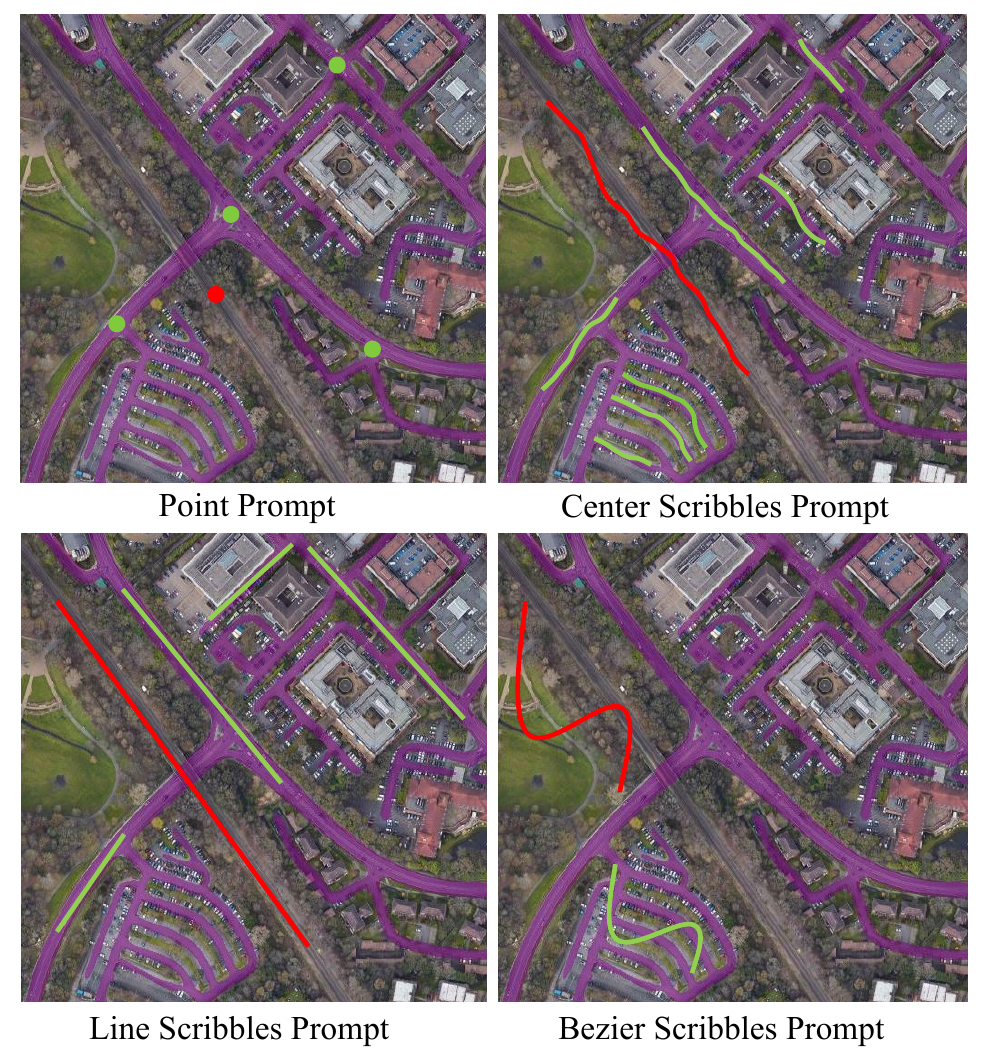}
    \captionsetup{font=small,skip=2pt}
    \caption{Simulated scribbles and clicks. Positive interactions (\textcolor{green}{green}) are simulated on the segmentation label, while negative interactions (\textcolor{red}{red}) are simulated on the background. Scribble thickness is enlarged for visual clarity.}
    \label{fig:prompt-type}
\end{figure}

\textbf{Point Interactions.}
A representative point $x^* \in V$ is sampled using a center-biased distance transform. The sampling probability is defined as $\displaystyle P(x) = \frac{\exp\left(\alpha \cdot E(x)\right)}{\sum_{z \in V} \exp\left(\alpha \cdot E(z)\right)}$, where $E(x)$ denotes the normalized euclidean distance transform within $V$, and $\alpha \in [1, 10]$ controls the degree of center bias. The selected point $x^*$ is rasterized as a soft disk and encoded into the positive or negative channel of the prompt map $P_n$.

\textbf{Scribble Interactions.}
Inspired by~\cite{wong2024scribbleprompt}, to simulate more expressive user corrections, we generate scribbles with varied geometry and structure. Specifically, we consider three types:

\begin{itemize}[leftmargin=8pt, topsep=2pt, itemsep=1pt, parsep=0pt]
    \item Center scribbles: Extracted from the skeleton of $V$ and truncated to reflect partial annotations.
    \item Line scribbles: Straight lines connecting two random points within $V$, emulating rapid strokes.
    \item Bezier scribbles: Smooth curves fitted through three control points sampled from $V$, mimicking natural drawing behavior.
\end{itemize}

All scribbles are rasterized using a fixed stroke width and perturbed by smooth displacement fields to simulate realistic variability. The resulting scribble masks are encoded into $P_n$, which is incrementally updated at each interaction round.

\begin{algorithm}[t]
\caption{Topo-Semantic Coupled Road Instantiation}
\label{alg:tscri}
\begin{algorithmic}
\Require Image $x$, mask $M$, prompt $q$, parameters $\theta$
\Ensure Instance mask $y$
\State $\mathcal{S} \gets \mathcal{F}_{\text{clean}}(x, M; \theta_{\text{clean}})$ \hfill $\triangleright$ structure refinement
\State $\mathcal{G} \gets \mathcal{F}_{\text{thin}}(\mathcal{S})$ \hfill $\triangleright$ centerline extraction
\State $\mathcal{A} \gets \mathcal{F}_{\text{attr}}(\mathcal{G})$ \hfill $\triangleright$ extract segment attributes
\State $\mathcal{C} \gets \text{Group}(\mathcal{A})$ \hfill $\triangleright$ segment grouping
\Statex \textbf{// Instance selection}
\State $s_q \gets \text{Score}(\mathcal{C}, q; \theta_{\text{sel}})$ \hfill $\triangleright$ learnable scoring
\State $e_q \gets \arg\max_{c_i \in \mathcal{C}} s_q[i]$ \hfill $\triangleright$ select top segment
\Statex \textbf{// Iterative region expansion}
\State Initialize region $\mathcal{R}_q^{(0)} \gets e_q$
\For{$t = 1$ to $T$}
    \State $\mathcal{R}_q^{(t)} \gets \Psi(\mathcal{R}_q^{(t-1)}, x; \theta_{\text{exp}})$ \hfill $\triangleright$ iterative expansion
\EndFor
\State $\mathcal{R}_q \gets \mathcal{R}_q^{(T)}$
\Statex \textbf{// Final refinement}
\State $y \gets \mathcal{F}_{\text{refine}}(\mathcal{R}_q, x; \theta_{\text{ref}})$ \hfill $\triangleright$ mask refinement
\State \Return $y$
\end{algorithmic}
\end{algorithm}

\begin{figure}[t]
    \centering
    \vspace{-3pt} 
    \includegraphics[width=0.48\textwidth]{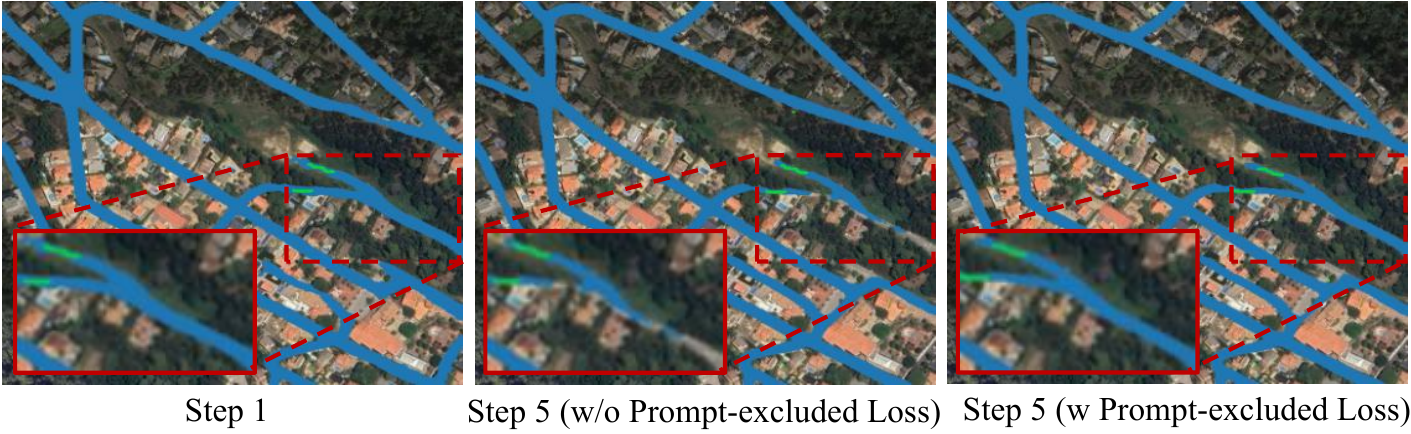}
    \captionsetup{font=small,skip=2pt}
    \caption{Visualization results before and after incorporating the Prompt-Excluded Skeleton Loss.}
    \label{fig:prompt-exclude}
    \vspace{-8pt} 
\end{figure}

\textbf{Expert-guided Prompt.}
To further guide the model toward uncertain or difficult regions (e.g., occluded or blurred road segments), we introduce an expert-guided sampling strategy based on disagreement between existing segmentation models and the ground truth.

Let $ \{M_j\}_{j=1}^{N} $ be a set of pre-trained segmentation models. For each training sample $x$, we compute the uncertainty mask $ \mathcal{U}(x) $ as the average absolute error across models:

\vspace{-1.2em}
\begin{equation}
\mathcal{U}(x) = \frac{1}{N} \sum_{j=1}^N \left| M_j(x) - y \right|.
\end{equation}
\vspace{-1.2em}

During prompt simulation, we define the prompt sampling probability map $ P(u = +1 \mid x) $ as:

\vspace{-1.2em}
\begin{equation}
P(u = +1 \mid x) = \frac{ \mathcal{U}(x)^{\beta} }{\sum_{z \in \Omega} \mathcal{U}(z)^{\beta}},
\end{equation}
\vspace{-1.2em}

where $ \Omega $ is the spatial domain and $ \beta > 1 $ controls sharpness. This formulation increases the likelihood of placing positive prompts in high-uncertainty regions, allowing the model to learn from ambiguous or hard-to-segment areas.

\subsection{Resolving Ambiguity}


In interactive road segmentation, user prompts guide model predictions. However, intent is often ambiguous. Some users may focus on major roads (e.g., highways), while others include multiple road types. This ambiguity is amplified in high-resolution satellite imagery, where road definitions vary across regions. Models trained on fixed labels often fail to generalize under such variation.

We alleviate this challenge with a topo-Semantic coupled road instantiation, detailed in Algorithm~\ref{alg:tscri}. Instead of directly predicting the mask, the model first regularizes the road structure using $\mathcal{F}_{\text{clean}}$, parameterized by $\theta_{\text{clean}} \subset \theta$. Centerlines are extracted via $\mathcal{F}_{\text{thin}}$, and segment-level attributes are computed through $\mathcal{F}_{\text{attr}}$. These segments are grouped, and a prompt-conditioned selector $\text{Score}(\cdot, q; \theta_{\text{sel}})$ ranks the candidates based on relevance to the input prompt.

During training, the model learns to resolve ambiguity by aligning prompt signals with the structured representation. The top-ranked segment is progressively expanded using $\Psi(\cdot; \theta_{\text{exp}})$ and refined via $\mathcal{F}_{\text{refine}}(\cdot; \theta_{\text{ref}})$ to produce the final instance mask. By deferring instantiation until after structural abstraction, the model achieves stronger generalization across regions and prompt variations.






\subsection{Loss Function}
Interactive segmentation benefits from user-provided prompts, but we observe a performance degradation during multi-step interactions, as shown in Fig.~\ref{fig:prompt-exclude}. Later prompts may overwrite initially correct regions, often leaving only sparse traces like thin lines. This reflects interactive degradation, where the model fails to maintain structural consistency across interaction steps.

\textbf{Prompt-excluded Skeleton Loss.} To address this, we adapt skeleton-based recall loss~\cite{kirchhoff2024skeleton} to preserve thin, elongated structures prone to being lost. However, applying it globally may cause overfitting on prompt regions, which are already well supervised. We therefore propose a prompt-excluded skeleton loss, which applies skeleton-based supervision only to non-prompt areas. Formally, let $\hat{y}_i$ and $y_i$ denote the predicted and ground-truth values at pixel $i$, and let $\mathcal{M}_n$ be a binary mask indicating the prompt region at step $n$ ($\mathcal{M}_n[i]=1$ if pixel $i$ is covered by a prompt). We define a prompt-free mask $\bar{\mathcal{M}}_n = 1 - \mathcal{M}_n$ to restrict loss computation to non-prompt regions.

\begin{figure*}[t]
    \vspace{-1em}
    \centering
    \includegraphics[width=\textwidth]{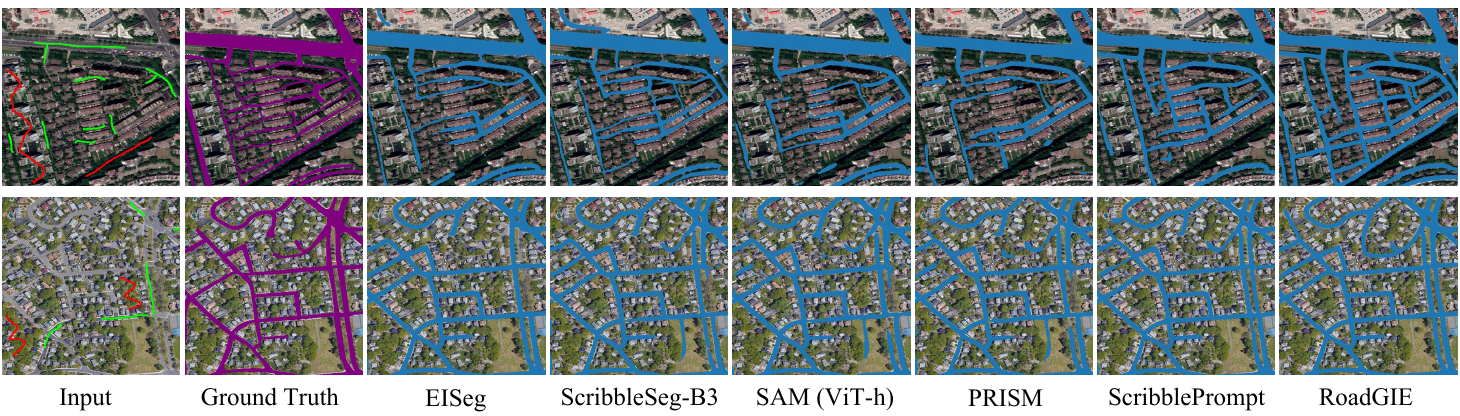} 
    \caption{Example predictions.}
    \label{fig:visual_results}
    \vspace{-1em}
\end{figure*}

The total training loss is defined as:

\vspace{-1em}
\begin{equation}
\begin{aligned}
\mathcal{L}_{\text{total}} =
&\underbrace{
-\sum_i \alpha (1 - \hat{y}_i)^\gamma y_i \log(\hat{y}_i)
}_{\text{Focal Loss}}
-
\underbrace{
\frac{2 \sum_i \hat{y}_i y_i + \epsilon}
{\sum_i (\hat{y}_i + y_i) + \epsilon}
}_{\text{Soft Dice Loss}} \\[4pt]
&-
\underbrace{
\frac{
  \sum_i \bar{\mathcal{M}}_n[i] \cdot \hat{y}_i \cdot \text{Skel}(y_i) + \epsilon
}{
  2 \sum_i \bar{\mathcal{M}}_n[i] \cdot \text{Skel}(y_i) + \epsilon
}
}_{\text{Prompt-excluded Skeleton Loss}}
\end{aligned}
\end{equation}
\vspace{-1em}

This formulation encourages the model to concentrate learning on uncertain or under-annotated regions, while mitigating redundancy and overfitting in areas already constrained by user prompts.
\section{Experiments}
\label{results}

\subsection{Settings}
\label{settings}

\textbf{Datasets.} We use the full \texttt{WorldRoadSeg-360K} dataset for training. The LSRV dataset serves as the test set. Additionally, the Global-Scale dataset and the combined others in Table~\ref{tab:dataset} (Baseline dataset) are used as baselines to validate the performance gain from training on \texttt{WorldRoadSeg-360K}.

\textbf{Evaluation Metrics.} Following~\cite{zhang2024graphmorph}, we adopt six evaluation metrics: Dice, Recall, clDice, APLS, $\beta0$, and $\beta1$. clDice combines Dice with connectivity consistency, making it suitable for tubular structures. APLS measures shortest-path similarity between predicted and ground-truth graphs, reflecting path-level connectivity. $\beta0$ counts connected components, indicating object-level topology, and $\beta1$ counts loops or holes, reflecting high-order topology.

\textbf{Implementation Details.} During training, each batch runs five interaction rounds using 1–3 visual prompts per round. These settings reflect typical user interaction patterns. Data augmentation includes rotation, flipping, contrast/brightness adjustment, and Gaussian blur. Training uses bf16 precision, AdamW optimizer, an initial learning rate of 0.0003, and a cosine schedule. Experiments run on 4×NVIDIA 3090 GPUs (24GB).

\subsection{Comparison with SOTAs}

Table~\ref{tab:magicbrush} compares the performance of different models using point and scribble prompts. Baseline dataset refers to the combined set of multiple road segmentation datasets. \texttt{RoadGIE} consistently outperforms all other models across both datasets. On Other dataset and \texttt{WorldRoadSeg-360k}, it surpasses the second-best model, ScribblePrompt, by 1.6 and 2.6 Dice points, respectively. Notably, no competing model achieves an improvement greater than 0.8 on both datasets. Fig.~\ref{fig:three_plots} (left) evaluates model performance at each interaction step, where \texttt{RoadGIE} consistently achieves the best results and starts with the highest initial Dice score.

\subsection{Ablation Study}

\begin{table}[t]
\vspace{-3pt}
\centering
\caption{\scriptsize Comparison of different models using different datasets.}
\label{tab:magicbrush}
\definecolor{Gray}{gray}{0.9}
\scriptsize
\setlength{\tabcolsep}{3.3pt} 
\renewcommand{\arraystretch}{1.05} 
\begin{tabular}{ccccc}
\toprule
\multirow{2}{*}{\textbf{Method}} & \multicolumn{2}{c}{\textbf{Baseline dataset}} & \multicolumn{2}{c}{\textbf{WorldRoadSeg-360K}} \\
\cmidrule(lr){2-3} \cmidrule(lr){4-5}
& Dice↑ & APLS↑ & Dice↑ & APLS↑ \\
\midrule
EISeg               & 0.701 & 0.511 & 0.706 & 0.515 \\
ScribbleSeg-B0 & 0.766 & 0.560 & 0.785 & 0.578 \\
ScribbleSeg-B3 & 0.761 & 0.556 & 0.788 & 0.580 \\
SAM (ViT-b)   & 0.719 & 0.522 & 0.737 & 0.539 \\
SAM (ViT-h)   & 0.738 & 0.539 & 0.756 & 0.553 \\
PRISM-2D              & 0.622 & 0.463 & 0.643 & 0.481 \\
PRISM-2D-Lite         & 0.656 & 0.489 & 0.669 & 0.496 \\
ScribblePrompt & 0.791 & 0.584 & 0.809 & 0.592 \\
\rowcolor{Gray}
\textbf{RoadGIE} & \textbf{0.807} & \textbf{0.593} & \textbf{0.835} & \textbf{0.620} \\
\bottomrule
\end{tabular}
\end{table}

\begin{table}[t]
\vspace{-5pt} 
\centering
\caption{Performance of models pretrained on different datasets and evaluated on the same test set. Best results are highlighted in bold.}
\label{tab:diffdataset}
\definecolor{Gray}{gray}{0.9}

\resizebox{\columnwidth}{!}{%
\begin{tabular}{ccccccc}
\toprule
\textbf{Pretrained dataset} & \textbf{Dice}↑ & \textbf{Recall}↑ & \textbf{clDice}↑ & \textbf{APLS}↑ & \textbf{\bm{$\beta_0$}}↓ & \textbf{\bm{$\beta_1$}}↓ \\
\midrule
Global-Scale        & 0.686          & 0.605          & 0.783          & 0.512          & 13.582          & 37.886 \\
Baseline dataset     & 0.807          & 0.897          & 0.869          & 0.593          & 8.150            & 3.061 \\
\cellcolor{Gray}{WorldRoadSeg-360K}   & \cellcolor{Gray}{\textbf{0.835}}  & \cellcolor{Gray}{\textbf{0.934}} & \cellcolor{Gray}{\textbf{0.905}} & \cellcolor{Gray}{\textbf{0.620}} & \cellcolor{Gray}{\textbf{5.823}}   & \cellcolor{Gray}{\textbf{2.752}} \\
\bottomrule
\end{tabular}%
}
\vspace{-5pt}
\end{table}

\textbf{WorldRoadSeg-360K.} To assess the effectiveness of the proposed dataset in improving model generalization, we trained models on three different datasets while keeping the test set fixed to LSRV. All metrics in Table~\ref{tab:diffdataset} are computed after 5 interaction steps. As shown in Table~\ref{tab:diffdataset}, training with \texttt{WorldRoadSeg-360k} achieves the best performance across all metrics. This demonstrates that \texttt{WorldRoadSeg-360k} can serve as a universal pretraining dataset for road segmentation tasks, effectively improving downstream performance.

\begin{figure}[t]
    \centering
    \includegraphics[width=0.48\textwidth]{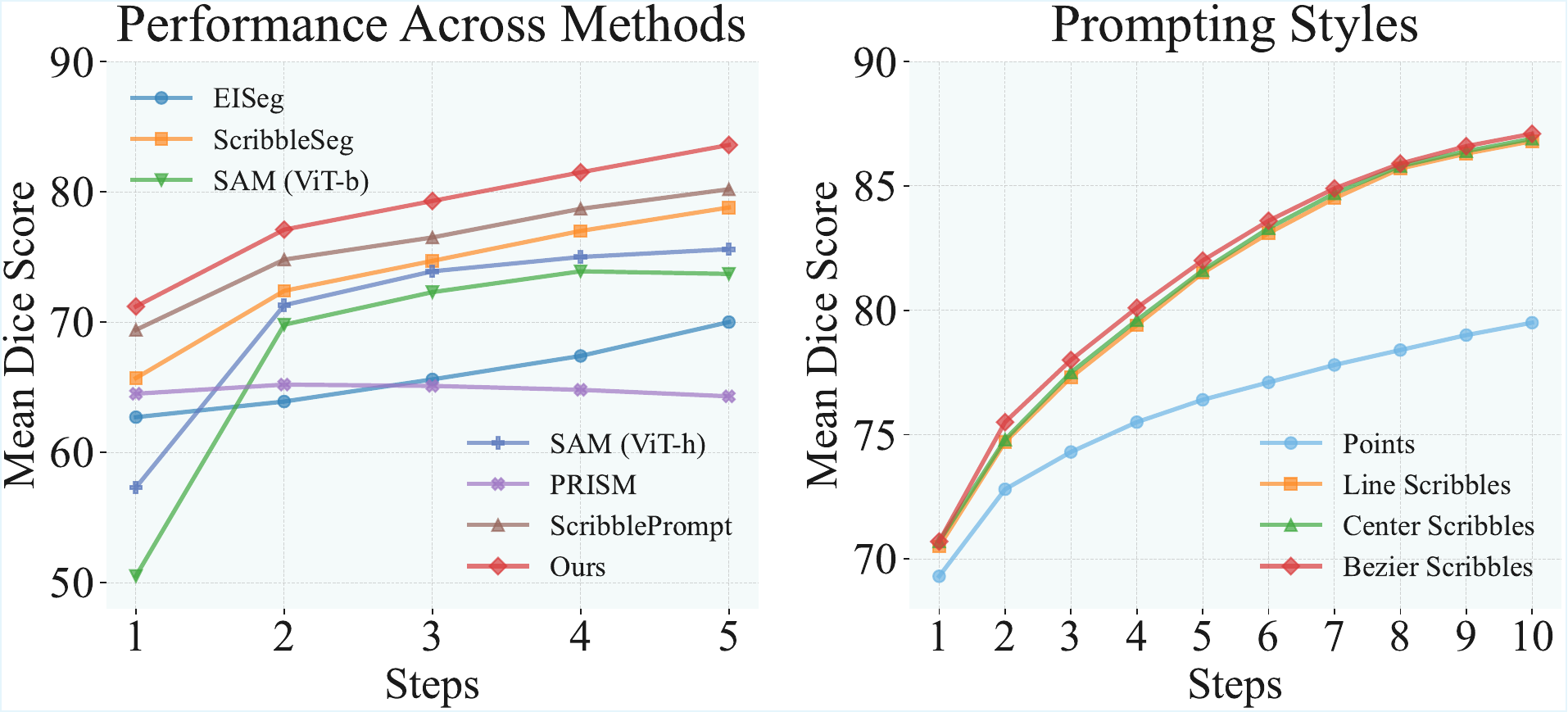}
    \caption{Interactive Performance. The left plot compares the performance of different methods; the right plot shows the performance of prompting types within our method.}
    \label{fig:three_plots}
\end{figure}

\textbf{Prompting Types.} We compare multiple prompting styles of \texttt{RoadGIE}. As shown in Fig.~\ref{fig:three_plots} (right), Bezier scribbles perform best, reaching a Dice of 87.1 after 10 iterations. The other two scribble types perform slightly worse, possibly due to limited guidance from regular scribbles. Points perform the weakest, with a Dice score below 80 after 10 iterations, confirming their limitation in high-aspect-ratio road segmentation tasks, where the model struggles to extract connectivity priors from sparse points.


\textbf{Expert-guided Prompt.} To highlight EG-Prompt’s gains, we reported per-round metrics in our earlier response, as shown in Table~\ref{tab:egprompt}: at round 5, it outperforms the baseline by 2.7\% in Dice and 2.9\% in APLS. Since annotators tackle harder regions in later rounds, such improvements underscore EG-Prompt’s strength in handling challenging samples. Fig.~\ref{fig:pipeline} further illustrates this trend.

\begin{table}[t]
\centering
\caption{Performance across five interaction rounds with and without EG-Prompt.}
\label{tab:egprompt}
\resizebox{0.47\textwidth}{!}{%
\begin{tabular}{l|c|ccccc}
\toprule
\textbf{EG-Prompt} & \textbf{Metric} & \textbf{Step 1} & \textbf{Step 2} & \textbf{Step 3} & \textbf{Step 4} & \textbf{Step 5} \\
\midrule
\multirow{2}{*}{\textbf{w/o.}} & Dice & 77.0 & 77.9 & 80.8 & 82.9 & 84.9 \\
                               & APLS & 56.8 & 57.2 & 59.9 & 61.5 & 63.7 \\
\midrule
\multirow{2}{*}{\textbf{w/.}} & Dice & \textbf{77.3} {\scriptsize \textbf{(+0.3)}} & \textbf{79.1} {\scriptsize \textbf{(+1.2)}} & \textbf{83.0} {\scriptsize \textbf{(+2.2)}} & \textbf{85.0} {\scriptsize \textbf{(+2.1)}} & \textbf{87.6} {\scriptsize \textbf{(+2.7)}} \\
                              & APLS & \textbf{57.0} {\scriptsize \textbf{(+0.2)}} & \textbf{58.5} {\scriptsize \textbf{(+1.3)}} & \textbf{61.7} {\scriptsize \textbf{(+1.8)}} & \textbf{64.2} {\scriptsize \textbf{(+2.7)}} & \textbf{66.6} {\scriptsize \textbf{(+2.9)}} \\
\bottomrule
\end{tabular}
}
\end{table}

\textbf{Topo-Semantic Coupled Road Instantiation.} We include visualizations of instantiated roads and interaction-time heatmaps in the supplementary material. Results show stronger activation in same-class road regions and improved connectivity in quantitative metrics, confirming the effectiveness of the topo-semantic instantiation for interactive road extraction.

\textbf{Prompt-excluded Skeleton Loss.} As shown in Table~\ref{tab:lossa}, Focal Loss and Dice Loss, although effective under class imbalance, are not designed to capture structural dependencies. Due to their localized supervision, applying prompt exclusion to these losses may weaken discrimination and destabilize training. In contrast, Skeleton-based Recall Loss operates at the structural level and evaluates the completeness of elongated objects such as roads. The prompt-exclusion mechanism naturally complements this objective by directing learning toward road-level contextual geometry rather than prompt-localized cues.

\begin{table}[t]
\centering
\caption{Loss strategies.}
\label{tab:lossa}
\definecolor{Gray}{gray}{0.9}
\resizebox{0.43\textwidth}{!}{%
\begin{tabular}{cccc|cc}
\toprule
\multicolumn{3}{c}{\textbf{Prompt-exclude Strategy}} & & \multicolumn{2}{c}{\textbf{Mean of 5 Steps}} \\
\midrule
\textbf{Focal} & \textbf{Dice} & \textbf{Skeleton-based Recall} & & \textbf{Dice} & \textbf{APLS} \\
\midrule
 & & & & 0.818 & 0.603 \\
\checkmark & & & & 0.806 & 0.595 \\
& \checkmark & & & 0.823 & 0.609 \\
& & \checkmark & & \textbf{0.829} & \textbf{0.615} \\
\bottomrule
\end{tabular}
}
\end{table}

\subsection{User Study}

\textbf{Inference Performance.} To assess the practicality of \texttt{RoadGIE}, we annotate 100 randomly sampled images from \texttt{WorldRoadSeg-360K}. As shown in Fig.~\ref{fig:pipeline}, manual annotation yields a Dice of 0.827, whereas \texttt{RoadGIE} improves it to 0.885, matching expert-level performance. \texttt{RoadGIE} also improves efficiency: it requires only 15s per image (7 interactions on average), reducing annotation time by 79\% compared with the 73s of manual labeling. These results highlight \texttt{RoadGIE} as both accurate and highly annotation-efficient.


\begin{table}[t]
\vspace{-1em}
\centering
\captionsetup{font=footnotesize}
\caption{GPU runtime comparison.}
\label{tab:runtime}
\scriptsize
\resizebox{\linewidth}{!}{%
\begin{tabular}{cc|cc}
\toprule
\textbf{Model} & \textbf{Inference Time (ms)} &
\textbf{Model} & \textbf{Inference Time (ms)} \\
\midrule
SAM (ViT-b)     & 283.65 & ScribbleSeg      & 97.31 \\
PRISM           & 305.36 & ScribblePrompt   & \textbf{30.76} \\
EISeg           & 270.28 & Ours             & 39.52 \\
\bottomrule
\end{tabular}%
}
\vspace{-1em}
\end{table}

\textbf{Inference Runtime.} We measure inference time separately on a single Intel(R) Xeon(R) Silver 4214 CPU @ 2.20GHz and on an Nvidia RTX3090 GPU for a prediction with a random scribble input covering 512 pixels. As shown in Table~\ref{tab:runtime}, on a single GPU, \texttt{RoadGIE} requires 39.52 ms per prediction, enabling the model to be used even in low-resource environments.
\section{Conclusion}


We introduce \texttt{WorldRoadSeg-360k}, a globally distributed large-scale road segmentation dataset comprising 366,947 images from 38 countries and 223 cities. \texttt{WorldRoadSeg-360k} facilitates robust model training and evaluation, and includes an out-of-domain test set to benchmark cross-domain generalization. Based on this, we propose \texttt{RoadGIE}, a new interactive paradigm for road extraction. \texttt{RoadGIE} supports connectivity-aware interactions, integrating both click- and scribble-based prompts to better capture topological continuity with minimal user effort. To improve structural consistency and avoid degradation during iterative interaction, it uses expert-guided prompting and adapts skeleton-based recall loss to the interactive setting. We also propose a topo-semantic instantiation that reduces user ambiguity by grounding prompts in spatial and semantic context, improving interaction stability and accuracy.

\textbf{Limitations.} The models and datasets are built on remote sensing imagery with 0.8–1.1m spatial resolution, which may limit their generalizability to much higher-resolution scenarios. Due to GPU memory constraints, training is limited to six interaction steps. In complex scenes, inference may require more steps than seen during training, potentially affecting segmentation accuracy.

\section*{Acknowledgement}

This research was supported by the Fund of the National Natural Science Foundation of China (Grant No. 62576177), Shenzhen Science and Technology Program (QNXMB20250701090801002, JCYJ20250604184027034, JCYJ20240813114237048), Guangdong Basic and Applied Basic Research Foundation (2026A1515011435), the Fundamental Research Funds for the Central Universities 070-63253222, and the Tianjin Key Laboratory of Visual Computing and Intelligent Perception (VCIP). Computation is supported by the Supercomputing Center of Nankai University (NKSC).  “Science and Technology Yongjiang 2035” key technology breakthrough plan project (2025Z053), Chinese government-guided local science and technology development fund projects (scientific and technological achievement transfer and transformation projects) (254Z0102G), National Science Fund of China under Grant No. 62361166670.

{
    \small
    \bibliographystyle{ieeenat_fullname}
    \bibliography{main}
}


\end{document}